\documentclass[conference]{IEEEtran}
\IEEEoverridecommandlockouts
\usepackage{cite}
\usepackage{amsmath,amssymb,amsfonts}
\usepackage{algorithmic}
\usepackage{graphicx}
\usepackage{textcomp}
\usepackage{xcolor}
\usepackage{balance}
\usepackage{threeparttable}
\usepackage{multirow}
\def\BibTeX{{\rm B\kern-.05em{\sc i\kern-.025em b}\kern-.08em
    T\kern-.1667em\lower.7ex\hbox{E}\kern-.125emX}}
\begin{document}

\title{RanAT4BIE: Random Adversarial Training for Biomedical Information Extraction\\
}


\author{
    \IEEEauthorblockN{Jian Chen$^{a,1}$, Shengyi Lv$^{a,1}$, Leilei Su$^{b,*}$\thanks{*Corresponding author.}}
    \IEEEauthorblockA{$^a$ Department of Data Science and Big Data Technology, Hainan University, Haikou 570228, China\thanks{ \textsuperscript{1}These authors contributed equally to this work.}}
    \IEEEauthorblockA{$^b$ Department of Mathematics, Hainan University, Haikou 570228, China}
    \IEEEauthorblockA{jian.chen.mail.cn@gmail.com, lsy0104@hainanu.edu.cn, leileisu225@gmail.com}
}

\maketitle

\begin{abstract}
We introduce random adversarial training (RAT), a novel framework successfully applied to biomedical information extraction (BioIE) tasks. Building on PubMedBERT as the foundational architecture, our study first validates the effectiveness of conventional adversarial training in enhancing pre-trained language models' performance on BioIE tasks. While adversarial training yields significant improvements across various performance metrics, it also introduces considerable computational overhead. To address this limitation, we propose RAT as an efficiency solution for biomedical information extraction. This framework strategically integrates random sampling mechanisms with adversarial training principles, achieving dual objectives: enhanced model generalization and robustness while significantly reducing computational costs. Through comprehensive evaluations, RAT demonstrates superior performance compared to baseline models in BioIE tasks. The results highlight RAT's potential as a transformative framework for biomedical natural language processing, offering a balanced solution to the model performance and computational efficiency.

\end{abstract}


\section{Introduction}
Adversarial training was initially conceptualized as a methodology for enhancing the robustness of deep learning models \cite{1DBLP:journals/corr/GoodfellowSS14}. This method is characterized by the deliberate introduction of controlled perturbations to training samples, thereby augmenting model resilience to input variations. While the efficacy of adversarial training in improving model robustness has been extensively demonstrated in computer vision applications, substantial evidence suggests that it may compromise model generalization capabilities \cite{2DBLP:conf/iclr/MadryMSTV18,3DBLP:conf/cvpr/XieWMYH19,4DBLP:conf/nips/ShafahiNG0DSDTG19}.

In previous work, Ali Shafahi et al. \cite{4DBLP:conf/nips/ShafahiNG0DSDTG19} documented significant performance degradation when implementing adversarial training methods. Specifically, the PGD method resulted in performance decreases of 7.76\% and 18.97\% on CIFAR-10 and CIFAR-100 datasets, respectively, while FreeAT exhibited declines of 9.05\% and 16.71\%. Intriguingly, contrary patterns have been observed in natural language processing, where adversarial training has been demonstrated to enhance both model generalization and robustness in downstream tasks \cite{5DBLP:conf/iclr/MiyatoDG17,6DBLP:conf/acl/ChengJM19}.

Empirical evidence presented by Miyato et al. \cite{5DBLP:conf/iclr/MiyatoDG17} revealed performance improvements through FGM-based adversarial training across multiple NLP classification datasets: IMDB \cite{7DBLP:conf/acl/MaasDPHNP11}, Elec \cite{8DBLP:conf/nips/JohnsonZ15}, RCV1 \cite{9DBLP:journals/jmlr/LewisYRL04}, Rotten Tomatoes \cite{10DBLP:conf/acl/PangL05}, and DBpedia \cite{11DBLP:journals/semweb/LehmannIJJKMHMK15}, with improvements of 1.18\%, 0.63\%, 0.28\%, 1.1\%, and 0.11\% respectively. Furthermore, Cheng et al. \cite{6DBLP:conf/acl/ChengJM19} demonstrated BLEU score improvements of 2.8\% and 1.6\% in Chinese-English and English-German translation tasks, respectively. Previous research has shown that the application of adversarial training has primarily focused on enhancing the security of machine learning systems across various domains, including autonomous driving \cite{12DBLP:conf/eccv/XiaoDLYLS18}, face recognition \cite{13DBLP:conf/cvpr/ChenZSLW22}, malware detection \cite{14DBLP:journals/eaai/ShaukatLV22}, and copyright monitoring \cite{15DBLP:conf/icml/SaadatpanahSG20}.

This study focuses on examining the impact of adversarial training on pre-trained language models' generalizability and its potential contributions to BioIE. Our findings demonstrate the efficacy of adversarial training in enhancing BERT model performance in BioIE tasks. A novel methodology, termed random adversarial training (RAT), has been proposed. Following the method of Miyato et al. \cite{5DBLP:conf/iclr/MiyatoDG17}, adversarial perturbations are applied to word embeddings, targeting the continuous rather than discrete sentence space. The implementation incorporates a Bernoulli distribution-inspired randomized adversarial training mechanism, whereby adversarial samples are stochastically selected during each training iteration, the specific architecture is shown in Figure 1. Our approach has been demonstrated to optimize training efficiency while simultaneously enhancing the pre-trained model's generalization capabilities.

\begin{figure*}
\centering
\includegraphics[width=0.9\textwidth]{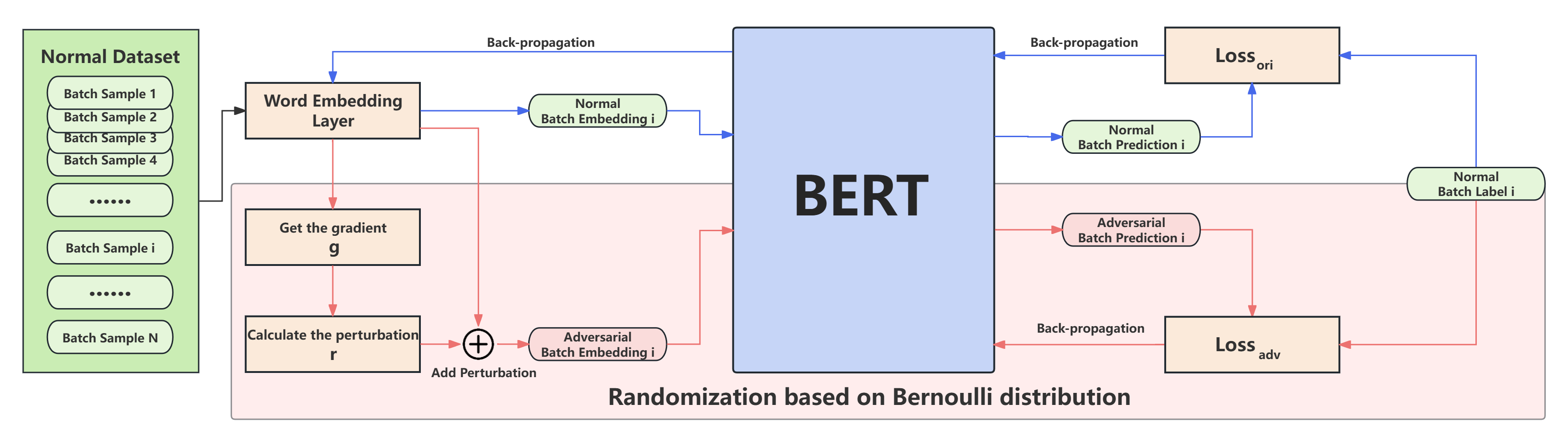}
\caption{Overall framework of our method.}
\end{figure*}

\section{Adversarial training}

\subsection{Basic Concepts}

The concept of adversarial samples was first established by Szegedy et al. \cite{16DBLP:journals/corr/SzegedyZSBEGF13}, who demonstrated that minimal perturbations to original samples could induce classification errors in neural networks. Subsequently, Goodfellow et al. \cite{1DBLP:journals/corr/GoodfellowSS14} introduced adversarial training and empirically demonstrated in computer vision applications that while model robustness was enhanced, generalization capabilities were compromised. Notably, this trade-off phenomenon has been observed to manifest differently in natural language processing, where both model robustness and generalization can be simultaneously improved \cite{5DBLP:conf/iclr/MiyatoDG17,6DBLP:conf/acl/ChengJM19}.

The conceptualization of adversarial samples, originally developed in computer vision, necessitates that adversarial examples maintain close proximity to their original counterparts. This is typically achieved through the introduction of perturbations in continuous space, ensuring that while the model can be deceived, human perception remains unaffected. The implementation involves various adversarial training methodologies, including FGM \cite{5DBLP:conf/iclr/MiyatoDG17}, PDG \cite{20DBLP:conf/emnlp/TranHHYTLP20}, FreeLB \cite{17DBLP:conf/iclr/ZhuCGSGL20}, and SMART \cite{18DBLP:conf/acl/JiangHCLGZ20}, to generate adversarial samples. These samples, in conjunction with original training data, are utilized to train deep learning models until sufficient resistance to adversarial attacks is achieved, thereby enhancing model robustness.

While natural language processing has adopted the principle of introducing minor perturbations in continuous space, the research focus has predominantly shifted toward improving model generalizability in downstream tasks through adversarial training. Although the theoretical foundation for this enhancement in NLP model generalizability remains incomplete, a notable perspective proposed by Aghajanyan et al. \cite{19DBLP:conf/iclr/AghajanyanSGGZG21} suggests that improvements in downstream model generalizability can be attributed to enhanced pre-trained model robustness. This hypothesis posits that fine-tuning for specific downstream tasks can lead to degradation of generic representations in pre-trained models, a phenomenon termed representation collapse. But, by implementing adversarial training to mitigate this representation collapse, pre-trained model robustness can be enhanced, subsequently facilitating improved model generalization across downstream tasks.

\subsection{Foundation framework}

The exceptional capacity of deep learning models to capture and represent input data patterns is manifested through their powerful fitting capabilities. However, this sensitivity to input features implies that when minimal perturbations are introduced to an original sample x, significant alterations in the resultant representations can occur, potentially leading to erroneous model predictions. Such deliberately perturbed instances, termed adversarial samples $x_adv$, can be mathematically expressed in relation to the perturbation r as:

\begin{equation}
\centering
x_{a d v}=x+r \text { s.t. }\|r\|_{p} \leq \epsilon
\label{eq.1}
\end{equation}

where $\epsilon$ is the perturbation magnitude and $\|\cdot\|_{p}$ is the $l_{p}$ paradigm, $p \in\{1,2, \infty\}$.

The fundamental principle of adversarial training lies in its utilization of adversarial samples for model training, thereby enhancing the model's resistance to potential attacks. Given a specified perturbation magnitude $\epsilon$, the adversarial training process aims to identify an optimal perturbation $r^*$ that maximizes the loss function. Consequently, the optimal perturbation $r^*$ can be mathematically formulated as:

\begin{equation}
\centering
r^{*}=\underset{\|r\|_{p} \leq \epsilon}{\arg \max } \mathcal{L}(f(x+r ; \theta), y)
\label{eq.2}
\end{equation}

where x is the original sample, y is the label, $\mathcal{L}(\cdot)$ is the loss function, $f(\cdot;\theta)$ is the neural network, and $\theta$ is a parameter in the neural network. Let $\mathcal{B}(r, \epsilon)=\left\{r \mid\|r\|_{p} \leq \epsilon\right\}$, then the above equation is specified as follows: given a loss function $\mathcal{L}(f(x+r ; \theta), y)$, find a perturbation $r^{*} \in \mathcal{B}(r, \epsilon)$ such that for all $r$ in $B$ there is $\mathcal{L}\left(f\left(x+r^{*} ; \theta\right), y\right) \geq \mathcal{L}(f(x+r ; \theta), y)$, where $B$ is the constraint set of the variable $r$, and the variable $r^*$ is the optimal solution in constraint set $B$.

By feeding the adversarial sample $x_{adv}=x+r^{*}$ into the model, the model parameter $\epsilon$ can be updated by minimizing the expected risk:
\begin{equation}
\centering
\theta^{*}=\arg \min _{\theta} \mathbb{E}_{(x, y) \sim \mathcal{D}}[\mathcal{L}(f(x+r ; \theta), y)]
\label{eq.3}
\end{equation}

Madry et al. \cite{2DBLP:conf/iclr/MadryMSTV18} summarized the above adversarial training methods into a min-max formulation from an optimization perspective, constructing adversarial training as an optimization problem:

\begin{equation}
\centering
\min _{\theta} \mathbb{E}_{(x, y) \sim D}\left[\max _{r \in \mathcal{B}(r, \epsilon)} \mathcal{L}(f(x+r ; \theta), y)\right]
\label{eq.4}
\end{equation}

This formulation characterizes adversarial training as a dual-objective methodology that simultaneously encompasses both offensive and defensive mechanisms. The inner optimization layer facilitates the generation of aggressive adversarial samples through loss maximization, while the outer optimization layer implements expected risk minimization for model parameter updates. This bidirectional method effectively integrates attack generation and defense reinforcement within a unified framework.

\subsection{Fast gradient sign method (FGSM) and fast gradient method (FGM)}

Goodfellow et al. \cite{1DBLP:journals/corr/GoodfellowSS14} introduced the Fast Gradient Sign Method (FGSM), a pioneering single-step adversarial training methodology initially conceived to enhance machine learning system security. This seminal work demonstrated the efficacy of adversarial training in significantly improving model robustness. FGSM represents the earliest implementation of gradient-based adversarial training, utilizing backpropagation gradients for perturbation computation. The mathematical formulation for generating adversarial samples under this framework can be expressed as:

\begin{equation}
\centering
x_{a d v}=x+\epsilon \cdot \operatorname{sign}\left(\nabla_{x} \mathcal{L}(f(x ; \theta), y)\right)
\label{eq.5}
\end{equation}
where $x$ is the original sample, $y$ is the label, $\epsilon$ is the perturbation amplitude, $sign(\cdot)$ is the sign function, $\nabla_{(\cdot)}$ is the gradient computed over the loss, $\mathcal{L}(\cdot)$ is the loss function, $f(\cdot;\theta)$ is the neural network, and $\epsilon$ is a parameter in the neural network. 

In this study, we followed the references of \cite{20DBLP:conf/emnlp/TranHHYTLP20} and \cite{21DBLP:journals/tcss/AlsmadiANAAKA23} and refer to the adversarial training method proposed by Miyato et al. \cite{5DBLP:conf/iclr/MiyatoDG17} as FGM, which was one of the early works to migrate adversarial training from the field of computer vision to the field of natural language processing. The adversarial sample generation formula for the FGM method is as follows:

\begin{equation}
\centering
x_{a d v}=x+\epsilon \cdot \frac{\nabla_{x} \mathcal{L}(f(x ; \theta), y)}{\left\|\nabla_{x} \mathcal{L}(f(x ; \theta), y)\right\|_{2}}
\label{eq.6}
\end{equation}
where the perturbations are no more computed using the $sign(\cdot)$ function, but instead the gradient is scaled by the $l_{2}$ paradigm.

The objective function for adversarial training consists of the standard training loss function and the adversarial training loss function as follows:
\begin{equation}
\centering
\mathcal{L}_{F G M}=\mathcal{L}(f(x ; \theta), y)+\mathcal{L}\left(f\left(x_{a d v} ; \theta\right), y\right)
\label{eq.7}
\end{equation}

\subsubsection{Projected gradient descent (PGD)}

Madry et al. \cite{2DBLP:conf/iclr/MadryMSTV18} advanced the field by introducing the Projected Gradient Descent (PGD) method, which represents a significant evolution in adversarial training methodology. In contrast to the single-step method of FGSM, PGD implements an iterative optimization framework, substantially enhancing the adversarial training process. The mathematical formulation for generating adversarial samples under the PGD framework can be expressed as:

\begin{equation}
\centering
r^{(t)}=\alpha \cdot \operatorname{sign}\left(\nabla_{x_{adv}^{(t-1)}} \mathcal{L}\left(f\left(x_{adv}^{(t-1)} ; \theta\right), y\right)\right)
\label{eq.8}
\end{equation}
 
\begin{equation}
\centering
x_{adv}^{(t)}=\Pi_{\left|x_{adv}^{(t-1)}-x\right| \leq \epsilon}\left\{x_{adv}^{(t-1)}+r^{(t)}\right\}
\label{eq.9}
\end{equation}

where $\epsilon$ is the perturbation magnitude, $\alpha$  is the perturbation step, usually $\alpha=\frac{\epsilon}{T}$, $\Pi\{\cdot\}$ is the projection function, $\nabla_{(\cdot)}$ is the gradient computed over the loss, $\mathcal{L}(\cdot)$ is the loss function, $f(\cdot;\theta)$ is the neural network, and $\epsilon$ is the parameter in the neural network. Special attention should be paid to $\left|x_{a d v}^{(t-1)}-x\right| \leq \epsilon$ as a constraint on the projection function, which theoretically maps the adversarial sample back into a hypersphere of radius $x+\epsilon$ when the absolute value of the difference between the adversarial sample and the original sample (i.e., the perturbation value) is larger than the perturbation magnitude.

In this study, we followed Miyato et al.'s method of replacing the $sign(\cdot)$ function using the $l_{p}$ paradigm. \cite{5DBLP:conf/iclr/MiyatoDG17}The formula is as follows: 

\begin{equation}
\centering
r^{(t)}=\alpha \cdot \frac{\nabla_{x_{a d v}^{(t-1)}} \mathcal{L}\left(f\left(x_{a d v}^{(t-1)} ; \theta\right), y\right)}{\left\|\nabla_{x_{a d v}^{(t-1)}} \mathcal{L}\left(f\left(x_{a d v}^{(t-1)} ; \theta\right), y\right)\right\|_{p}}
\label{eq.10}
\end{equation}

In engineering practice, the iteration of the adversarial sample $x_{adv}$ can be realized by implementing only the iteration of the perturbation r. The following is an equivalent expression of the above equation:

\begin{equation}
\mathrm{r}^{\prime}=\alpha \cdot \frac{\nabla_{r^{(t-1)}} \mathcal{L}\left(f\left(x+r^{(t-1)} ; \theta\right), y\right)}{\left\|\nabla_{r^{(t-1)}} \mathcal{L}\left(f\left(x+r^{(t-1)} ; \theta\right), y\right)\right\|_{p}}
\label{eq.11}
\end{equation}

\begin{equation}
r^{(t)}=\Pi_{\left\|r^{(t-1)}\right\| p \leq \epsilon}\left\{r^{(t-1)}+r^{\prime}\right\}
\label{eq.12}
\end{equation}

\begin{equation}
x_{a d v}^{(t)}=x+r^{(t)}
\label{eq.13}
\end{equation}

In this study, the projection function is set to:
\begin{equation}
\Pi_{\left\|r^{(t-1)}\right\|_{2} \leq \epsilon} \left\{r^{(t)}\right\}=\frac{\epsilon}{\left\|r^{(t)}\right\|_{2}} \cdot r^{(t)}
\label{eq.14}
\end{equation}

The objective function of PGD is as follows:
\begin{equation}
\mathcal{L}_{P G D}=\mathcal{L}(f(x ; \theta), y)+\mathcal{L}\left(f\left(x_{a d v}^{(t)} ; \theta\right), y\right)
\label{eq.15}
\end{equation}

\subsubsection{Free Large Batch Adversarial Training (FreeLB)}

Zhu et al. \cite{17DBLP:conf/iclr/ZhuCGSGL20} introduced Free Large-Batch (FreeLB), an enhanced multi-step adversarial training methodology specifically designed for natural language processing applications. FreeLB can be conceptualized as a refined iteration of the PGD framework, with a crucial architectural distinction: while PGD utilizes solely the final attack gradient for parameter updates, FreeLB implements an averaged gradient method across m attacks, enabling more comprehensive parameter optimization. While FreeLB maintains the same adversarial sample generation formulation as PGD, its distinctive objective function can be expressed as:

\begin{equation}
\mathcal{L}_{\text {FreeLB }}=\mathcal{L}(f(x ; \theta), y)+\frac{1}{T} \sum_{t=1}^{T} \mathcal{L}\left(f\left(x_{a d v}^{(t)} ; \theta\right), y\right)
\label{eq.16}
\end{equation}

\subsubsection{SMoothness inducing Adversarial Regularization (SMART)}

SMART, formally known as SMAR3T2, was introduced by Jiang et al. \cite{18DBLP:conf/acl/JiangHCLGZ20}. The framework comprises two principal components: SMoothness-inducing Adversarial Regularization and BRegman pRoximal poinT opTimization. While the former addresses adversarial training aspects, the latter focuses on optimization algorithms. For the scope of this study, we concentrate specifically on the adversarial training component. Notably, SMART employs the same adversarial sample generation mechanism as PGD.

The distinguishing characteristic of SMART lies in its treatment of adversarial loss as a regularization term. The algorithm implements a dual-phase method: utilizing KL-loss for the initial k-1 steps, followed by symmetric KL-loss in the final k-th step. Analogous to PGD, SMART exclusively employs the adversarial loss from the terminal step for gradient updates. However, SMART represents a significant departure from conventional adversarial training algorithms. While traditional approaches target the adversarial sample directly and employ identical loss functions (such as cross-entropy loss) for both standard and adversarial training, SMART introduces a novel paradigm. Its adversarial objective focuses on maximizing the disparity between pre- and post-perturbation outputs. Rather than measuring the divergence between the adversarial sample's predicted probability distribution and the true distribution, SMART quantifies the discrepancy in model outputs before and after adversarial sample application. This can be formally expressed as:

\begin{equation}
\min _{\theta} \mathcal{F}(\theta)=\mathcal{L}(\theta)+\alpha \mathcal{R}(\theta)
\label{eq.17}
\end{equation}

where $\mathcal{F}(\cdot)$ is the objective function, $\mathcal{L}(\cdot)$ is the loss function for normal training, $\mathcal{R}(\cdot)$ is the loss function for adversarial training is viewed as a regularization term, and $\alpha$ is a hyperparameter that regulates the strength of regularization.

It is further given that when the loss function for adversarial training is in the form of KL-loss and symmetric KL-loss:

\begin{equation}
\mathcal{L}_{K L}(P \| Q)=\sum_{i} P(i) \log \frac{P(i)}{Q(i)}
\label{eq.18}
\end{equation}

\begin{equation}
\mathcal{L}_{\text {sym-KL }}(P, Q)=\mathcal{D}_{K L}(P \| Q)+\mathcal{D}_{K L}(Q \| P)
\label{eq.19}
\end{equation}

where $L_{KL}(\cdot)$ is the KL-loss, $L_{sym-KL} (P,Q)$ is the symmetric KL-loss, and $P(\cdot)$ and $Q(\cdot)$ are the probability distributions of the outputs before and after the maximization perturbation, respectively. 

The adversarial loss formulae involved in the gradient update are given:
\begin{equation}
\mathcal{R}(\theta)=\frac{1}{n} \sum_{i}^{n} \max _{\left\|x_{a d v}^{(t)}-x_{i}\right\|_{p} \leq \epsilon} \mathcal{L}_{s y m-K L}\left(f\left(x_{a d v}^{(t)} ; \theta\right), f\left(x_{i} ; \theta\right)\right)
\label{eq.20}
\end{equation}

\begin{figure*}
\centering
\includegraphics[scale=0.14]{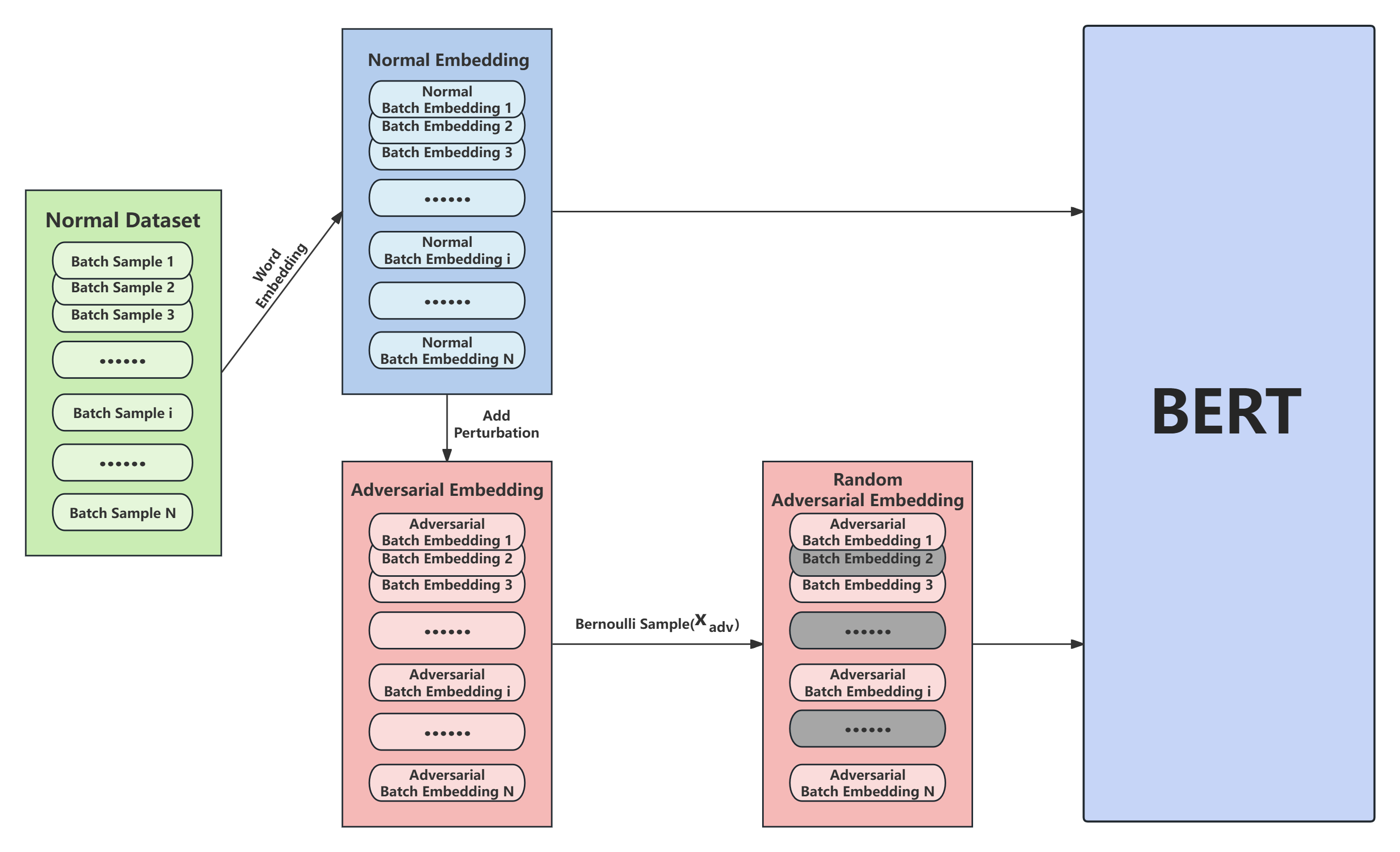}
\caption{Our adversarial example generation strategy.}
\end{figure*}

\section{Random Adversarial Training}

Adversarial training fundamentally operates by introducing controlled perturbations to training samples, thereby enhancing the model's resilience to input variations. These precisely calibrated perturbations are designed to induce suboptimal model performance, effectively stress-testing the system. The implementation typically involves established adversarial training methodologies, including FGM \cite{5DBLP:conf/iclr/MiyatoDG17}, PGD \cite{2DBLP:conf/iclr/MadryMSTV18}, FreeLB \cite{17DBLP:conf/iclr/ZhuCGSGL20}, and SMART \cite{18DBLP:conf/acl/JiangHCLGZ20}, to generate adversarial samples. These samples, in conjunction with the original training data, are utilized iteratively until the model demonstrates robust resistance to adversarial attacks.

Our proposed random adversarial training (RAT) framework (Figure 1) introduces a novel probabilistic perspective to this process. It conceptualizes the adversarial sample generation for each batch during every epoch as a sequence of independent Bernoulli trials, where each generation attempt represents a binary random event with consistent probability. Under this framework, the decision to proceed with adversarial sample generation is determined probabilistically before processing each batch. Upon successful generation, the framework executes standard adversarial training procedures with the generated samples. Conversely, when the generation event fails, the process is terminated, and the adversarial training phase is bypassed for that particular batch. Figure 2 details our adversarial example generation strategy.

Firstly, the definition of an adversarial sample is given:
\begin{equation}
x_{a d v}=x+r \text { s.t. }\|r\|_{p} \leq \epsilon
\label{eq.21}
\end{equation}

where r is the small perturbation, $\epsilon$ is the perturbation magnitude,  $\|\cdot\|_{p}$ is the $l_{p}$ paradigms and $\mathrm{p} \in\{2, \infty\}$.

Then, the definition of Bernoulli distribution is given:
\begin{equation}
P\{k\}=p^{k}(1-p)^{1-k} \text { s.t. } k=\{0,1\}
\label{eq.22}
\end{equation}

where p represents the probability of the event being successful and $k$ represents a random variable that represents success when $k = 1$ and conversely represents failure. We will denote the random variable satisfying the distribution as above as $k\sim Bernoulli(p)$.

Suppose that the set of original samples of a batch is $\text { Batch }^{i}=\left\{x_{t}\right\}_{t=1}^{\text {BatchSize}}$ and the set of generation events of the adversarial samples is $\mathcal{D}_{\text {event }}=\left\{\text { event }_{i}\right\}_{i=1}^{n / \text { BatchSize }}$. One can describe this process of generating adversarial samples following a Bernoulli distribution as:
\begin{equation}
\mathcal{D}_{\text{event}}=\left\{\text {event}_{1}, \ldots \text{event }_{n / \text {BatchSize}}\right\} \sim \operatorname{Bernoulli}(p)
\label{eq.23}
\end{equation}

\begin{equation}
\text {Batch}^{i}=\text {event}_{i} \times\left\{x_{adv}^{t}\right\}_{t=1}^{\text {BatchSize}}
\label{eq.24}
\end{equation}

where, $event_{i}$ denotes the $i\mathcal{-}th$ adversarial sample generation event, where the success or failure is determined stochastically. Upon successful realization, the framework proceeds with adversarial sample generation and subsequent training; conversely, the adversarial training phase is omitted. The collective set of generation events follows a Bernoulli distribution.

Subsequently, each batch comprising original samples x and their corresponding adversarial counterparts $x_{adv}$ is fed into the model for training. The model parameters $\theta$ are then optimized through the minimization of the expected risk, which can be expressed as:

\begin{equation}
\theta=\arg \min_{\theta} \mathbb{E}_{(x, y) \sim \mathcal{D}}\left[\mathcal{L}(f(x;\theta), y)+\mathcal{L}\left(f\left(x_{adv};\theta\right), y\right)\right]
\label{eq.25}
\end{equation}
where D is the original data distribution and $E[\cdot]$ is the expectation.

\section{Experiments and Analysis}
\subsection{Definition of Tasks}

\begin{table*}[htbp]
\renewcommand{\thetable}{\uppercase{i}}
  \centering
  \caption{Standard Training Performance v.s. Adversarial Training Performance.}
  \begin{threeparttable}
  \begin{tabular}{lcccccc}
    
    \hline  
    Method	& BC5-chem(\%) & BC5-disease(\%) &	BC2GM(\%) &	NCBI-disease(\%) &	CHEMPROT(\%) &	DDI(\%) \\
    \hline  
    BERT\cite{28DBLP:conf/naacl/DevlinCLT19}	&89.25 	&81.44 	&80.90 	&85.67 	&71.86 	&80.04 \\
    RoBERTa\cite{29liu2019roberta}	&89.43 	&80.65 	&80.90 	&86.62 	&72.98 	&79.52 \\
    BioBERT\cite{30DBLP:journals/bioinformatics/LeeYKKKSK20}	&92.85 	&84.70 	&83.82 	&\textbf{89.13} 	&76.14 	&80.88 \\
    SciBERT\cite{31DBLP:conf/emnlp/BeltagyLC19}	&92.49 	&84.54 	&83.36 	&88.10 	&75.00 	&81.22 \\
    ClinicalBERT\cite{32DBLP:journals/corr/abs-1904-03323}	&90.80 	&83.04 	&81.71 	&86.32 	&72.04 	&78.20 \\
    BlueBERT\cite{33DBLP:conf/bionlp/PengYL19}	&91.19 	&83.69 	&81.87 	&88.04 	&71.46 	&77.78 \\
    PubMedBERT\cite{34DBLP:journals/health/GuTCLULNGP22}	&93.33 	&85.62 	&84.52 	&87.82 	&77.24 	&82.36 \\

    \hline
    PubMedBERT+FGM &93.61 	&\textbf{85.96} 	&84.82 	&88.67 	&80.27 	&\textbf{83.20} \\
    PubMedBERT+PGD	&\textbf{93.68} 	&85.80 	&\textbf{84.95} 	&88.56 	&80.87 	&83.03 \\
    PubMedBERT+FreeLB &93.63 	&85.72 	&84.73 	&88.89 	&\textbf{81.13} 	&83.01 \\
    PubMedBERT+SMART &93.48 	&85.81 	&84.52 	&88.89 	&79.93 	&82.49 \\
\hline
    
  \end{tabular}
 \begin{tablenotes}    
        \footnotesize
        \item[1] The \textbf{bolded} items denotes the best performance.        
    \end{tablenotes}
\end{threeparttable}
\end{table*}

\subsubsection{BioNER Task Definition}
For the BioNER task will be given the dataset $D=\{X_1,X_2,…,X_N\}$ and the named entity E. The formalization is defined as follows: 
\begin{equation}
f\left(X_{i} ; \theta\right)=\left\{\left(E, I_{\mathrm{s}}^{1}, I_{e}^{1}\right), \ldots,\left(E, I_{\mathrm{s}}^{t}, I_{e}^{t}\right), \ldots,\left(E, I_{\mathrm{s}}^{T}, I_{e}^{T}\right)\right\}
\end{equation}

where $X_{i}=\left\{x_{1}, x_{2}, \ldots, x_{n}\right\} \in D$ is a vector, and $\left(E_{m}, I_{\mathrm{s}}^{t}, I_{e}^{t}\right)$ is the entity recognition triple for the BioNER task. 
In this study, we employ the Begin-Inside-Outside (BIO) tagging scheme for word-level annotation, effectively transforming the BioNER task into a multi-class classification problem at the word level\cite{chen2025multimodal}. For the word-level multi-class classification task will be given the dataset $D=\left\{X_{1}, X_{2}, \ldots, X_{N}\right\}$, the label set $Y=\left\{Y_{1}, Y_{2}, \ldots, Y_{N}\right\}$ and the set of label categories $C=\{B, I, O\}$. where $X_{i}=\left\{x_{1}, x_{2}, \ldots, x_{n}\right\} \in D$ is a sequence vector, $Y_{i}=\left\{y_{1}, y_{2}, \ldots, y_{n}\right\} \in Y \text { s.t. } y_{i} \in C$ is a label sequence. The goal of the word-level multi-category classification task is to categorize words from any sample $X_{i}=\left\{x_{1}, x_{2}, \ldots, x_{n}\right\}$, so the model is actually computing the probability $P\left(C_{k} \mid x_{i}\right)$:

\begin{equation}
\begin{aligned}
\operatorname{Softmax}&\left(f\left(\left\{x_{1}, \ldots, x_{n}\right\} ; \theta\right)\right)\\&=\left\{P\left(C_{k} \mid x_{1}\right), \ldots, P\left(C_{k} \mid x_{n}\right)\right\}
\end{aligned}
\end{equation}
And to get the predicted labels, simply pick the predicted label with the highest probability:

\begin{equation}
\begin{aligned}
\operatorname{argmax}&
\left(\left\{P\left(C_{k} \mid x_{1}\right), \ldots, P\left(C_{k} \mid x_{n}\right)\right\}\right)\}\\
&=\left\{\hat{y}_{1}, \ldots, \hat{y}_{n}\right\}
\end{aligned}
\end{equation}

Where $\left\{\hat{y}_{1},  \ldots, \hat{y}_{n}\right\}$ is the predicted label sequence.

\subsubsection{BioRE Task Definition}
For the BioRE task will be given a dataset $D={X_1,X_2,…,X_N}$ and a predefined set of entity relationships $R=\left\{R_{1}, R_{2}, \ldots, R_{K}\right\}$. The formalization is defined as follows:
\begin{equation}
f\left(X_{i} ; \theta\right)=\left\{\left(S_{1}, R_{k}, O_{1}\right), \ldots,\left(S_{t}, R_{k}, O_{t}\right), \ldots,\left(S_{T}, R_{k}, O_{T}\right)\right\}
\end{equation}

where $X_{i}=\left\{x_{1}, x_{2}, \ldots, x_{n}\right\} \in \mathrm{D}$ is a vector and $\left(S_{t}, R_{k}, O_{t}\right)$ is the entity-relationship triple for the BioRE task. 

Our study aligns with established methodologies in biomedical relationship extraction, specifically concentrating on relationship classification within the broader extraction framework. Our approach frames the relationship extraction task as a classification problem among identified entities. More precisely, the process involves pairwise matching of entities within each original sample $X_i$, followed by the decomposition into multiple relation classification instances $X_{RC-i}$. The formal definition of our relation extraction task can be concisely expressed as:

\begin{equation}
f\left(X_{R C-i} ; \theta\right)=(S, R, O)
\end{equation}

According to the formal definition above, this study regards the task as a sentence-level multi-class classification task. For the sentence-level multi-class classification task will be given the dataset $D=\left\{X_{R C-1}, X_{R C-2}, \ldots, X_{R C-N}\right\}$, the set of labeled categories $C=\left\{C_{1}, C_{2}, \ldots C_{K}\right\}$ and the label set $Y=\left\{Y_{1}, Y_{2}, \ldots, Y_{N}\right\} \in C$. The goal of the sentence-level multi-category classification task is to classify the relation $C_k$ of any relationally categorized sample $X_(RC-i)$, so the model is actually computing the probability $P\left(C_{k} \mid X_{R C-i}\right)$:

\begin{equation}
\operatorname{Softmax}\left(f\left(X_{R C-i} ; \theta\right)\right)=P\left(C_{k} \mid X_{R C-i}\right)
\end{equation}

And to get the predicted labels, simply pick the predicted label with the highest probability:
\begin{equation}
\operatorname{argmax}\left(P\left(C_{1} \mid X_{R C-i}\right), \ldots, P\left(C_{k} \mid X_{R C-i}\right)\right)=\hat{Y}
\end{equation}

\begin{table*}[tbp]
  \renewcommand{\thetable}{\uppercase{ii}}
  \centering
  \caption{Standard Training Performance v.s. Adversarial Training Performance}
  \begin{threeparttable}
  \begin{tabular}{clcccccc}
    
    \hline  
    Training & Method	&BC5-chem(\%)	&BC5-disease(\%)	&BC2GM(\%)	&NCBI-disease(\%)	&CHEMPROT(\%)	&DDI(\%) \\
    \hline 

    Standard	&baseline	&93.33 	&85.62 	&84.52 	&87.82 &77.24 	&82.36 \\
    \hline
        \multirow{5}{*}{AT}	
        &FGM	&93.61 	&85.96 	&84.82 	&88.67 	&79.21 	&83.20 \\ 
	&PGD	&93.68 	&85.80 	&84.95 	&88.56 	&79.42 	&83.03 \\
	&FreeLB	&93.63 	&85.72 	&84.73 	&88.89 	&79.25	&83.01 \\
	&SMART	&93.48 	&85.81 	&84.52 	&88.89 	&77.56 	&82.49 \\
	&Avg	&\textbf{93.60} 	&85.82 	&\textbf{84.76} 	&88.75 	&\textbf{78.86} 	&82.93 \\

    \hline
        \multirow{5}{*}{RAT}	
	&FGM	&93.64 	&85.93 	&84.82 	&88.79 	&79.17 	&83.75 \\
	&PGD	&93.63 	&85.82 	&84.84 	&89.00 	&79.20 	&83.52 \\
	&FreeLB	&93.68 	&85.80 	&84.79 	&88.89 	&79.17 	&83.34 \\
	&SMART	&93.43 	&85.88 	&84.54	&89.02 	&77.65 	&83.34 \\
	&Avg	&\textbf{93.60} 	&\textbf{85.86} 	&84.75 	&\textbf{88.93} 	&78.80 	&\textbf{83.39} \\

\hline

  \end{tabular}
    \begin{tablenotes}    
        \footnotesize
        \item[1] The Avg values indicate the average performance of all methods for a particular training framework.  
        \item[2] Bolded values indicate the best performance.        
    \end{tablenotes}
 \end{threeparttable}
\end{table*}

\subsection{Performance Experiments}

\subsubsection{Adversarial training performance analysis}

In this section, we evaluate the efficacy of adversarial training in enhancing model generalization capabilities across six prominent biomedical datasets: Bc5cdr-chem \cite{22DBLP:journals/biodb/LiSJSWLDMWL16}, Bc5cdr-disease \cite{22DBLP:journals/biodb/LiSJSWLDMWL16}, BC2GM \cite{23smith2008overview}, NCBI-disease \cite{24DBLP:conf/bionlp/CollierK04}, CHEMPROT \cite{25krallinger2017overview}, and DDI \cite{26DBLP:journals/jbi/Herrero-ZazoSMD13,27segura2013semeval}, using mainstream pre-trained models as benchmarks. Throughout our experimental framework, PubMedBERT serves as the backbone architecture, with micro-F1 score adopted as the primary evaluation metric.

The experimental results presented in Table \uppercase{i} demonstrate substantial performance improvements when implementing adversarial training with PubMedBERT \cite{34DBLP:journals/health/GuTCLULNGP22} as the foundation model. Specifically, we observe performance enhancements of 0.35\%, 0.34\%, 0.43\%, 1.07\%, 3.89\%, and 0.84\% across BC5-chem, BC5-disease, BC2GM, NCBI-disease, CHEMPROT, and DDI datasets, respectively. These consistent improvements across diverse datasets provide compelling evidence for the effectiveness of adversarial training in enhancing model generalization capabilities.

\subsubsection{Random Adversarial training performance analysis}
In this section of experiments, PubMedBERT is used as the baseline, and adversarial training and random adversarial training are applied to PubMedBERT, respectively.The goal of the random adversarial training is to improve the training efficiency of the model without decreasing the model generalizability. The success rate of adversarial attacks for random adversarial training in the experiments in this section is uniformly 50\%, i.e., compared with standard adversarial training, random adversarial training can reduce the number of attacks by 50\%, which in turn improves the training efficiency of the model.

Table \uppercase{ii} presents the results of the experiments comparing standard training, adversarial training (AT), and our proposed approach (RAT) across NER and RE datasets. The results show that both AT and RAT consistently outperform the standard training baseline across all datasets. For the BC5-chem dataset, both AT and RAT achieve an average performance of 93.60\%, which is a 0.27 percentage point improvement over the baseline of 93.33\%. In the case of BC5-disease, AT and RAT show slight improvements, with RAT (85.86\%) performing marginally better than AT (85.82\%), both surpassing the baseline (85.62\%). In the BC2GM dataset, AT and RAT demonstrate similar improvements, with average scores of 84.76\% and 84.75\% respectively, compared to the baseline of 84.52\%. The NCBI-disease dataset shows more substantial gains, with RAT achieving the highest average score of 88.93\%, followed closely by AT at 88.75\%, both significantly outperforming the baseline of 87.82\%. For the CHEMPROT dataset, both AT and RAT show considerable improvements over the baseline (77.24\%), with AT achieving an average of 78.86\% and RAT slightly lower at 78.80\%. Lastly, on the DDI dataset, RAT demonstrates the best performance with an average of 83.39\%, followed by AT at 82.93\%, both surpassing the baseline of 82.36\%.

It's worth noting that while AT and RAT show similar overall performance, RAT achieves these results with 50\% fewer adversarial attacks during training, thus drastically improving training efficiency. The results suggest that random adversarial training can maintain or even slightly improve model performance while reducing computational costs compared to standard adversarial training. Furthermore, among the different adversarial attack methods (FGM, PGD, FreeLB, and SMART), no single method consistently outperforms the others across all datasets. This suggests that the choice of adversarial attack method may depend on the specific characteristics of each dataset or task.

\subsection{Efficiency Analysis}
This section presents a theoretical analysis of the computational efficiency gains achieved through random adversarial training compared to conventional adversarial training approaches. In deep learning implementations, the computational cost ratio between backward propagation (BP) and forward propagation (FP) typically maintains a 1:2 relationship, where one BP operation equals the computational cost of two FP operations. Consequently, we standardize all computational costs in terms of FP operations (xFP).

In Table \uppercase{iii}, S denotes the number of attack iterations in multi-step adversarial training. From a theoretical standpoint, S can approach positive infinity, and this asymptotic behavior is incorporated into our Cost Reduction Rate (CRR) calculations. The comparative analysis presented in Table \uppercase{iii} delineates the training costs (expressed in xFP) and CRR across three training paradigms: standard training, conventional adversarial training, and random adversarial training. The results demonstrate that random adversarial training significantly reduces computational overhead, achieving a remarkable CRR of up to 50\% compared to conventional adversarial training methods.

\begin{table}[tbp]
\renewcommand{\thetable}{\uppercase{iii}}
  \centering
  \caption{Standard Training Performance v.s. Adversarial Training Performance}
  \begin{tabular}{clcccc}
    \hline
    Training &Method	&FP	&BP	&xFP	&CRR \\
    \hline

    Standard &baseline	&1	&1	&3	&- \\
    \hline
        \multirow{4}{*}{AT}	
        &FGM	&2	&2	&6	&- \\
        &PGD	&1+S	&1+S	&3+3S	&-\\
        &FreeLB	&1+S	&1+S	&3+3S	&-\\
        &SMART	&1+S	&1+S	&3+3S	&-\\
    \hline
        \multirow{4}{*}{RAT} 
        &FGM	&1.5	&1.5	&4.5	&25\% \\
	&PGD	&1+0.5S	&1+0.5S	&3+1.5S	&50\% \\
	&FreeLB	&1+0.5S	&1+0.5S	&3+1.5S	&50\% \\
	&SMART	&1+0.5S	&1+0.5S	&3+1.5S	&50\% \\

    \hline
  \end{tabular}
\end{table}

\section{Conclusion}
In this study, we proposed and evaluated a novel random adversarial training approach for biomedical information extraction tasks. The experiments demonstrated that RAT can maintain or even slightly improve model performance compared to standard adversarial training, while significantly reducing computational costs by up to 50\%. This highlights the effectiveness of RAT in enhancing both the robustness and efficiency of pre-trained language models for domain-specific applications, and opens up new possibilities for the practical application of advanced NLP technologies in biomedical research and clinical settings.


\balance
\bibliographystyle{unsrt}
\bibliography{ref}

\end{document}